\documentclass[runningheads]{llncs}
\usepackage{graphicx}
\usepackage{todonotes}
\usepackage{hyperref}
\begin{document}
\title{Towards explainable meta-learning}
\author{Katarzyna Woźnica\inst{1} \and
Przemysław Biecek\inst{1} }

\institute{Warsaw University of Technology \\
\email{katarzyna.woznica.dokt@pw.edu.pl}\\
}
\maketitle              % typeset the header of the contribution
\begin{abstract}
Meta-learning is a field that aims at discovering how different machine learning algorithms perform on a wide range of predictive tasks.  Such knowledge speeds up the hyperparameter tuning or feature engineering. With the use of surrogate models various aspects of the predictive task such as meta-features, landmarker models e.t.c. are used to predict the expected performance. State of the art approaches are focused on searching for the best meta-model but do not explain how these different aspects contribute to its performance. However, to build a new generation of meta-models we need a deeper understanding of the importance and effect of meta-features on the model tunability. In this paper, we propose techniques developed for eXplainable Artificial Intelligence (XAI) to examine and extract knowledge from black-box surrogate models. To our knowledge, this is the first paper that shows how post-hoc explainability can be used to improve the meta-learning.

\end{abstract}

\begin{figure*}[t]
   \centering\includegraphics[width=\textwidth]{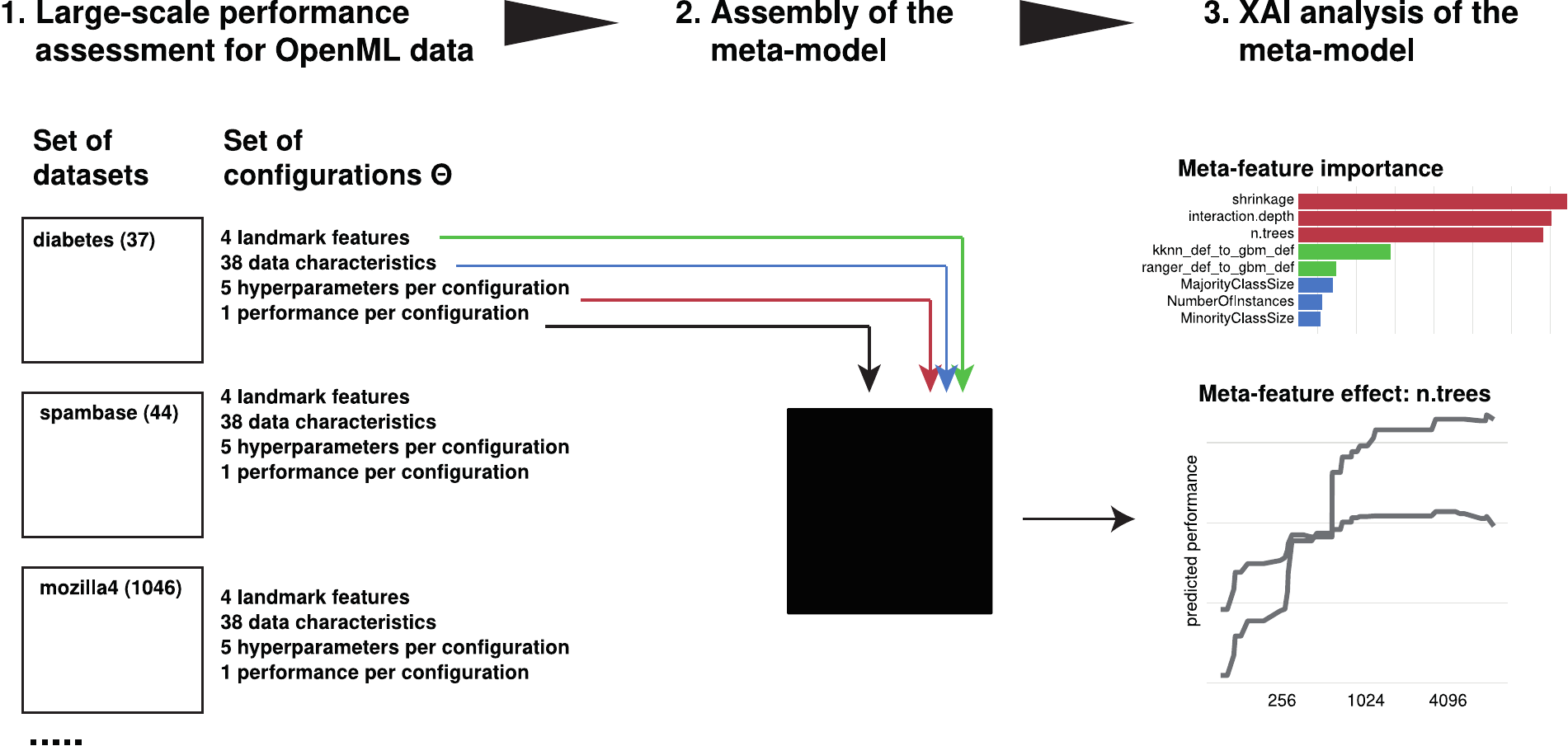}
   \caption{The proposed process of the explainable meta-model exploration. First, we gather meta-features about selected data sets from OpenML repository. Then we calculate model performance on these data sets for selected configurations of hyperparameters. Second, we assemble a black-box surrogate meta-model. Third, we use XAI techniques to extract information about the relative importance of meta-features and their marginal responses.}
   \label{graphical_abstract}
 \end{figure*}
 
\section{Introduction}
\label{section:introduction}

Meta-learning, or learning to learn, is an area of machine learning focused on extracting and transferring knowledge about tuned models from one predictive problem to another. 
It may increase the effectiveness of AutoML systems by automating feature selection, hyperparameters tuning and optimization of model performance \cite{Vanschoren2019}. By using previously acquired experience, meta-learning allows to narrow the domain of promising solutions for a new prediction problem what results in quicker and more computationally efficient model development.  

Meta-learning is a very broad term, difficult to define, and is perceived differently depending on the research area \cite{Vanschoren2019,hospedales2020meta}. The main feature characterizing meta-learning is what aspect of the learning process is being leveraged and transferred. In the case of deep learning, a common object of interest is finding initial values of weights for some neural network architecture such that after several  gradients steps to obtain accurate predictions \cite{finn2017model,finn2018probabilistic}. For tabular data, machine learning hyperparameters configuration may be considered as transferred meta-representation since it often turns out that state-of-the-art results are obtained by classical models such as ensembles of tree based models \cite{friedman2001greedy,chen2016xgboost,ke2017lightgbm,dorogush2018catboost}. Then the space of potential configurations has several to a dozen dimensions  so practitioners need resource-effective methods to develop well-performing ML models and need guidelines to make the right choices. In this work we narrow the scope of meta-learning to recommendation of hyperparameters settings to initializing machine learning algorithms for tabular data.

So far, researchers have presented  diverse approaches to plug-in results from prior experiments into model optimization for novel predictive problem. Most of them are based on the principle of interaction between two submodules~\cite{jomaa2021dataset2vec}. The first is a meta-feature extractor, which compresses the data set into an vector of predefined features of predictive problems. For  example this compression may  consist of statistical summaries for a single variable or groups of variables, or it may be based on the evaluation of set of baseline ML algorithms, so called landmarker models~\cite{LeiteSelectingDatasets}. The second submodule is a meta-learner model that uses meta-features to predict a meta-target. Meta-models for tabular data may assess the most promising hyperparameters configuration~\cite{rendell1990empirical,king1995statlog}, the ranking of ones~\cite{pinto2017autobagging,lorena2018data} or the performance prediction for given hyperparameters settings~\cite{guerra2008predicting,reif2014automatic,davis2018annotative,probst2019tunability}. Very often, a similarity assessment between the training meta-data and independent test meta-data, is the component of rules of the meta-models. Distinct methods consolidating of measurable properties of data sets and model configurations have proven their benefit in optimization~\cite{feurer2015initializing,Wistuba2016SequentialTuning,wistuba2018scalable}.

The effectiveness of meta-learning is  highly dependent on the type and quality of meta-features~\cite{bilalli2017predictive}.  So it is crucial to consider a broad range of candidates~\cite{vilalta2004using}.
The primary sources of meta-features that feasibly predict model performance are data sets' characteristics. Conventional approach is to express whole data set as a vector of  simple engineered statistics. 
A comprehensive summary of these commonly used meta-features is provided in \cite{Vanschoren2019,Rivolli2018TowardsMeta-learning}. Autoencoders as meta-features extractors \cite{edwards2016towards,hewitt2018variational}  avoid the problem of a priori defining meta-features but they are limited to the same schema of meta-data sets. More versatility is offered by an extractor Dataset2Vec based on hierarchical modelling  \cite{jomaa2021dataset2vec}.
In addition to defining features through the internal data structure, we can also take into account how difficult a given prediction problem is for a certain portfolio of machine learning models. This is a concept of landmarking models  introduced  in~\cite{pfahringer2000meta}. It extends the list of meta-features by the relative performance of some predefined models, so-called landmarkers. 
Except for data set's properties, good meta-models take into account the tunability of particular algorithms with respect to selected hyperparameters \cite{probst2019tunability}. 
For that reason, part of the meta-feature space is detailed model configuration - user-determined hyperparameters.

Another subdomain of machine learning that has been developing intensively in recent times is eXplainable Artificial Intelligence (XAI). With the growing demand for the highly engineered models there is  apparent  necessity of investigation black-box algorithms and identification the most important aspects affecting the model operation.  There are the reservoir of XAI methods which address the justifications of a model’s predictions for a single instance as well as the data set-level exploration \cite{Biecek,molnar2019}. So far these exploratory techniques are applied to the enhancement of ML model but this kind of inference would amplify the meta-learning generalization ability.  The transparency of the algorithm structure helps to identify which explanatory variables are included in the predictive process and how changes in their distribution affect the model. What is more, the effectiveness of meta-models is strongly dependent on the chosen set of meta-targets and the approach to modeling the relationship between them and the meta-response. So similar to classical machine learning approaches, it can be a trial-and-error process requiring several iterations.

In this work, we integrate these two promising  directions (see Figure~\ref{graphical_abstract}).  We show how to use XAI techniques to extract knowledge and validate of meta-models used in meta-learning. We focus on the specific architecture of meta-models which enables application of exploratory techniques. Using them we extract essential informative properties of meta-features. 
We present an example for meta-learning based on OpenML100 data sets, but the presented approach can be applied to any meta-model trained for particular domain set of problems. To our knowledge, this is the first paper that combines these two areas of machine learning.

\section{Meta-learning frameworks}

In this section we briefly summarize the two most common  approaches to predicting model performance for given hyperparameter configuration. As we mention, the first step in meta-learning is extracting meta-features characterizing the data sets. According to the way of utilizing these meta-features we may define following  classes f meta-learning frameworks. 

\begin{enumerate}

 \item \textbf{Sequential surrogate meta-models} leverage meta-features to assess the similarity between meta-train data sets and new task and then use these measures as weights to configuration transfer. Meta-learner does not use meta-features explicitly, they are compressed to similarity measure so it is more challenging to understand meta-features impact.
 
 To propose new hyperparameters that have the potential for good model performance on the new dataset, extensions of Sequential Model Based Optimization (SMBO) are used. The MI-SMBO procedure~\cite{feurer2015initializing} uses the best predicted hyperparameters from similar data sets to select the initial points to optimization for novel data. There are proposed two definitions of similarity between meta-features vectors: the $p$-norm of the difference between the data sets’ meta-features and metric that reflects how similar the data sets are with respect to the performance of different hyperparameters settings. The meta-features modification is utilized into a major step in optimization in a surrogate Gaussian Process (GP) for all tasks simultaneously \cite{yogatama2014efficient}. To model similarities between instances, they use a squared exponential kernel and the nearest neighbours kernel. This approach is extended to be more scalable  in \cite{wistuba2015learning,wistuba2018scalable}. They fit separate GP for each task in meta-data, then aggregate them into one using the Nadaraya Watson kernel.
    
    \item \textbf{Black-box surrogate meta-models} directly predicts model performance for given hyperparameters settings and given data set described with  vector of meta-features. Meta-learner is actually a regression model and meta-data has tabular format - every data set is represented as row with a sequence of meta characteristics~\cite{brazdil2008metalearning}.
    Firstly, this generic approach is proposed in~\cite{metal}. They provide knn-ranking method to rank candidate models. In  \cite{vilalta2004using} authors also consider these type of approach in meta-model framework and suggest the potential source of meta-features. In \cite{reif2014automatic} authors use SVM model as meta-regressor and utilize broad scope of meta-features: statistical and information theory measures, model based features and landmarkers. For every machine learning algorithms they train meta-learner independently because of different hyperparameters configuration. Similar meta-features are applied  by~\cite{davis2018annotative} but meta-learner is a Multilayer Perceptron. Authors of \cite{probst2019tunability} extend the idea of surrogate benchmarks and  train surrogate regression to map hyperparameter configuration  to model performance.

\end{enumerate}

These both perspectives on  meta-learning show empirically that transferring knowledge from independent tasks is beneficial with respect to yielding from that meta-framework the optimal hyperparameter  or more effective warm start points optimization for the new task. The sequential approach requires the validation of each step separately and it is more difficult to explore interactions between meta-features and hyperparameter configurations. On the other hand, the black box meta-model is very similar to the classic machine learning approach. It therefore offers great opportunities for analogous model exploration methods. Therefore in this paper we focus on the latter approach and we show how eXplainable Artificial Intelligence methods can be used to audit or extract knowledge from black-box meta-model.

\section{The Meta-OpenML100 surrogate model}
\label{section:meta_learning}

In the next section, we present a universal approach to an exploration of the black-box meta-model. Both the choice of data for which model evaluation and hyperparameters transfer was carried out and the choice of algorithm for meta-learner should be considered as illustrative. Similarly to previous researches, this meta-model works on classification problems from OpenML100 benchmark  \cite{bischl2017openml}. We use a meta-model engineering methodology analogous to \cite{reif2014automatic,davis2018annotative,probst2019tunability} and employ gradient boosting algorithm as meta-learner. In this section, we describe how this meta-model is built (Figure~\ref{graphical_abstract}).

We built meta-model based on all predictive tasks for binary classification in OpenML100 suite, see the list of these tasks in Table \ref{tableOptimalMetaFeatures}. According to \cite{vilalta2004using} recommendations, for each task, we calculate the following meta-features: four landmarkers performance for baseline models and 38 statistical properties of the underlying data set. The meta-model is built to predict the performance of gradient boosting model parametrized with five hyperparameters. Finally, we train a meta-learner, also a gradient boosting model, to predict the performance of a model with selected hyperparameters on a data set with following meta-properties.

This black-box surrogate model is called Meta-OpenML100 in the rest of this article. In following subsections, we provide a detailed description of its components.

 \subsection{Predictive tasks and their meta-features}
 
 Out of all tasks in OpenML100 suite \cite{bischl2017openml}  we select only these for binary classification. The suite provides 61 meta-characteristics for these tasks \cite{bilalli2017predictive}, some correspond to properties of continuous variables, some for categorical variables and some for data sets with mixed variables. Here we limit ourselves only to data sets with all continuous variables, i.e. 20 data sets listed in Table \ref{tableOptimalMetaFeatures}.  For these data sets, we use 38 available statistical and information-theoretic properties. 

% latex table generated in R 3.6.2 by xtable 1.8-4 package
% Thu Feb  6 14:17:30 2020
\begin{table*}[t!h]
\caption{Meta-features for selected data sets from OpenML repository. Four landmarks (relative performance to default gbm model) and five hyperparameters for gbm model are presented. Data sets' characteristics are omitted for the brevity. Only optimal hyperparameters are listed in last columns for the corresponding data set. }
\label{tableOptimalMetaFeatures}
\vskip 0.15in
\begin{center}
% \begin{small}
\begin{scriptsize}
\begin{sc}
\begin{tabular}{p{30mm}|p{9mm}p{9mm}p{9mm}p{9mm}p{9mm}|p{8mm}p{9mm}p{9mm}p{9mm}}
  \hline
 & \multicolumn{5}{c|}{Hyperparameters for GBM model}  & \multicolumn{4}{c}{Landmarks}  \\ 
data set (id) & shrink. & inter. \newline depth & n.trees & bag \newline fract. & min. node & knn  & glmnet  & ranger   & random forest   \\ 
  \hline
  diabetes (37) & 0.00 &   4 & 1480 & 0.69 &   7 & 1.10 & 2.25 & 2.36 & 2.30 \\ 
  spambase (44) & 0.04 &   5 & 1414 & 0.98 &  16 & 2.97 & 4.78 & 7.57 & 7.74 \\ 
  ada\_agnostic (1043) & 0.05 &   5 & 333 & 0.90 &   7 & 2.31 & 5.41 & 6.28 & 5.37 \\ 
  mozilla4 (1046) & 0.09 &   4 & 1567 & 0.54 &   7 & 1.71 & 0.39 & 2.62 & 2.84 \\ 
  pc4 (1049) & 0.01 &   5 & 2367 & 0.75 &  12 & 1.11 & 2.13 & 3.46 & 3.57 \\ 
  pc3 (1050) & 0.04 &   1 & 949 & 0.90 &  11 & 0.86 & 1.34 & 1.85 & 1.88 \\ 
  kc2 (1063) & 0.00 &   2 & 273 & 1.00 &  21 & 0.81 & 0.73 & 0.90 & 0.98 \\ 
  kc (1067) & 0.00 &   3 & 5630 & 0.26 &   3 & 0.23 & 1.16 & 1.49 & 1.66 \\ 
  pc1 (1068) & 0.00 &   4 & 6058 & 0.21 &  14 & 1.12 & 0.27 & 1.92 & 1.84 \\ 
  banknote authentication (1462) & 0.03 &   1 & 8429 & 0.52 &  12 & 6.47 & 4.99 & 5.56 & 6.02 \\ 
  blood transfusion service center (1464) & 0.01 &   1 & 394 & 0.21 &  10 & 0.64 & 1.23 & 0.86 & 0.71 \\ 
   climate model simulation crashes (1467) & 0.00 &   2 & 654 & 0.26 &  15 & 0.35 & 1.36 & 1.12 & 1.15 \\ 
   eeg-eye-state (1471) & 0.08 &   5 & 2604 & 0.28 &  14 & 2.48 & 0.93 & 3.34 & 4.16 \\ 
   hill-valley (1479) & 0.08 &   5 & 2604 & 0.28 &  14 & 1.78 & 0.43 & 2.04 & 2.24 \\ 
   madelon (1485) & 0.05 &   5 & 333 & 0.90 &   7 & 0.34 & 0.48 & 1.61 & 1.58 \\ 
   ozone-level-8hr (1487) & 0.00 &   3 & 8868 & 0.33 &  11 & 1.41 & 3.40 & 4.43 & 4.36 \\ 
   phoneme (1489) & 0.02 &   5 & 5107 & 0.60 &  18 & 4.89 & 2.90 & 7.12 & 8.13 \\ 
   qsar-biodeg (1494) & 0.05 &   5 & 333 & 0.90 &   7 & 4.16 & 5.34 & 6.74 & 6.81 \\ 
   wdbc (1510) & 0.04 &   1 & 949 & 0.90 &  11 & 1.56 & 0.63 & 1.88 & 1.91 \\ 
   wilt (1570) & 0.00 &   4 & 6058 & 0.21 &  14 & 2.73 & 4.80 & 7.15 & 7.28 \\ 
   \hline
\end{tabular}
\end{sc}
\end{scriptsize}
% \end{small}
\end{center}
\end{table*}

 \subsection{Landmarkers}

We use landmarkers-based meta-features to characterise predictive problems. As landmarkers models, we consider five machine learning algorithms with default hyperparameter configurations:\textit{generalized linear regression with regularization}, 
 \textit{gradient boosting},
\textit{k nearest neighbours}, 
 two random forest implementations: \textit{randomForest  and ranger}. 
Various models have been proposed as landmarkers \cite{balte2014meta} but one of the requirements is diversified architectures of algorithms capturing different variables interrelationship. We balance this diversity with the landmarkers'  computational complexity \cite{pfahringer2000meta}.

To evaluate their predictive power we apply 20 train/test split method and compute AUC scores for each split. These five algorithms are ranked according to methodology in section~\ref{Model_dataset predictive power}. Because we predict the performance of gradient boosting models we compute landmarkers as a ratio of models rankings (knn, glmnet, ranger and randomForest)  to a ranking of gbm model with the default configuration. In a result, we have four landmarkers meta-features.

 \subsection{Algorithms and hyperparameters space}
  \label{Algorithms_hyperparameters_space}

In this paper we explore the performance of gradient boosting classifiers (\textit{gbm}) to selection of  following hyperparameters: \textit{n.trees}, 
 \textit{interaction.depth},
 \textit{n.minobsinnode}, 
\textit{shrinkage}, 
 \textit{bag.fraction}. In order to do so, we sample 100 random configurations of hyperparameters in a similar way to \cite{probst2019tunability} to examine their influence gradient boosting model predictive power. Additionally, we add one special configuration - the default settings for the \texttt{gbm} library \cite{gbm}.

  \subsection{Estimated predictive power of selected configurations}
   \label{Model_dataset predictive power}

 For every combination of 20 data sets from OpenML100 suite and the 101 hyperparameters configurations, we try pre-specified 20 train and test data splits. Each model is fitted on each training subset and afterwards, AUC is computed on the test frame.  This way we obtain a meta-data set for the performance of 40400 configuration/data sets/split combinations. Because performance for different data sets takes values in different ranges we normalized these values by using ranks per data set. 
 The higher AUC, the higher position in the ranking. Ratings are scaled to [0,1] interval. Every configuration for each algorithm appears in list 20 times because of train-test splits. To aggregate this to one value for every model we  computed average rating for the model.

\subsection{Surrogate meta-model}

As meta-model we select gradient boosting algorithm with maximum interaction depth equals 10  i.e. that model may be rich in interactions between meta-features. This kind of algorithms has been already applied inside sequential-model based optimization for example in SMAC \cite{hutter2011sequential}. Tree-based algorithms are particularly well suited to handle high-dimensional and partially categorical input spaces. They are known for robustness and automated feature selection. 

 This surrogate meta-model configuration  is selected as the best one according to evaluation schema as follows. We apply one-data set-out cross-validation: every model is trained on 19 data sets and then is tested on the remaining data frame by calculating mean square error (MSE) and Spearman correlation between predicted rankings and actual meta-responses.

The meta-data set with all meta-features and fully reproducible code can be found in GitHub repository \url{https://github.com/woznicak/MetaFeaturesImpact}.

   \section{Explanatory analysis of Meta-OpenML100 model}
   \label{section:explainable_analysis}
   
 The black-box meta-model approach leads to the fact that even if it works effectively, we do not acquire new knowledge about how data set characteristics translates to  optimal choices for hyperparameters.
In this chapter we present the main result of this work - recommendations on how one can use selected XAI techniques \cite{Biecek,molnar2019} for the analysis of black-box meta-model. Each recommendation is complemented with an example for the Meta-OpenML100 meta-model developed in the previous section. Often the knowledge we draw from the model looks intuitive, but only thanks to presented an analysis we get a quantitative validation of our assumptions.

 The XAI  methods allow to analyse single meta-model. In Meta-OpenMl100 we use one-data set-out schema so we can explore the   constructions' each of these meta-models independently or extend this approach and aggregate feature importance across crossvalidation meta-models.

 \subsection{Meta-features importance}
 \label{subsection:features_importance}

    The meta models are built on various sets of meta-features determined for data sets. For complex meta-models it is difficult to determine which variables actually contribute to the model output.  This investigation is needed to identify presumptive noisy aspects and may be significant in deliberation to exclude these meta-features from new generations meta-models. The solution to this problem is to use model agnostic permutational feature importance, that assess how perturbations of a specific feature decrease model performance \cite{fisher2019all}.
    
\begin{figure}[h]
    \centering
    
    \includegraphics[width=0.9\textwidth ]{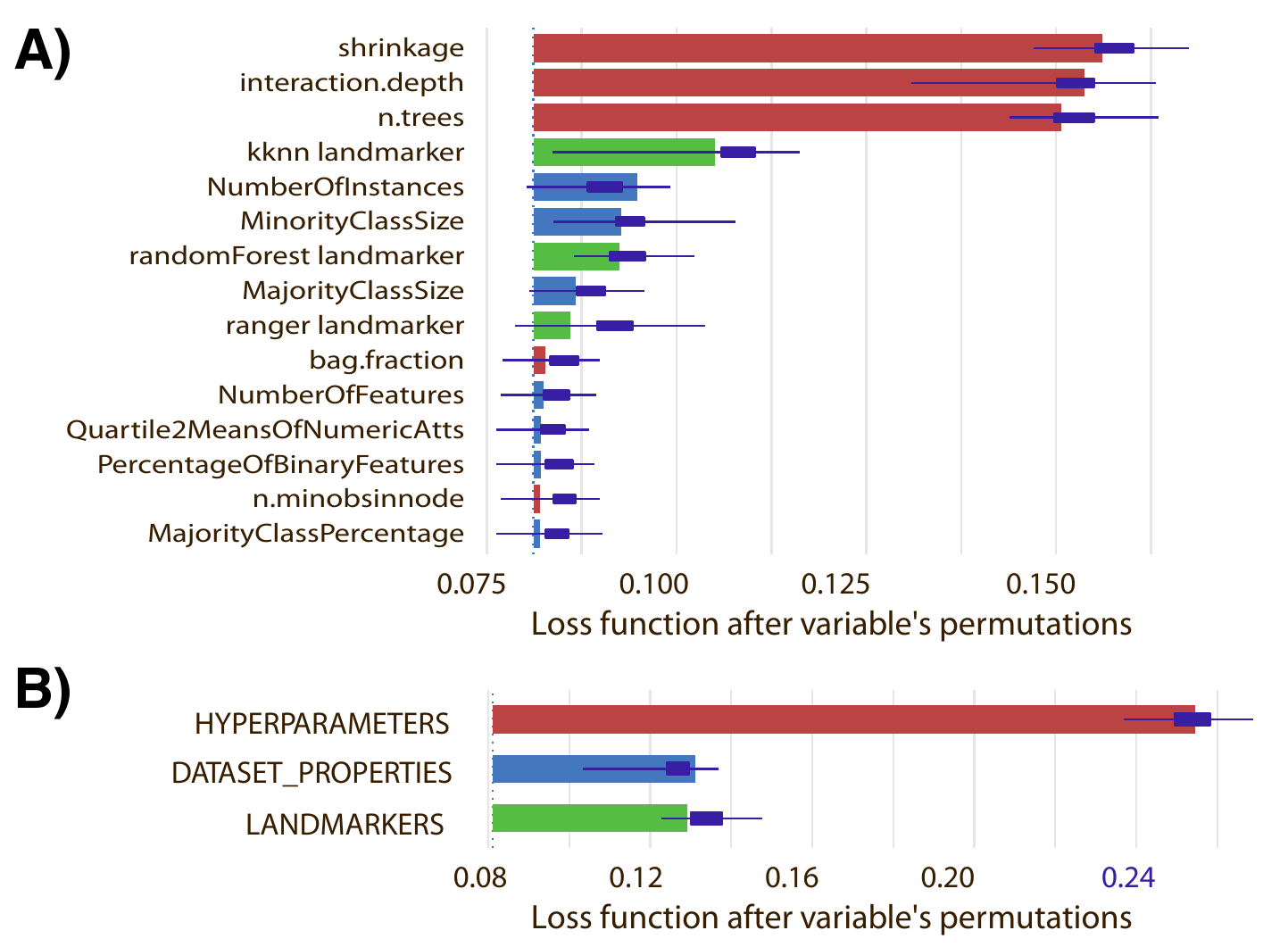}\vspace{-2mm}
    \caption{Panel A: Importance of the 15 top meta-features in GBM meta-model. Panel B: Cumulative importance of groups of meta-features.}
    \label{plot_top15_most_imp_feature_agg}

\end{figure}

      An example for Meta-OpenMl100 is presented in Figure \ref{plot_top15_most_imp_feature_agg}. We can easily read that the most important are hyperparameters followed by two landmarker features (\texttt{knn} and \texttt{randomForest}) and two meta-features (\texttt{NumberOfInstances} and \texttt{MinorityClassSize}).
      
      Considered meta-features form three groups because of a different approach to creating them. Figure \ref{plot_top15_most_imp_feature_agg}B presents the assessment of groups influence. As we see, the most important class of meta-features is hyperparameters and this conclusion is consistent with the importance measure for individual variables. Landmarkers and data set characteristics have similar dropout values.
      
\subsection{Meta-features interactions}
A good meta-model offers different optimal hyperparameters for different data sets  to solve the problem of no one optimal hyperparameters across tasks. This means that model detects interactions between variables, situations in which the values of one attribute (meta-characteristics) affect the effect of another attribute (hyperparameter). 
In general, identification of interactions is a difficult problem. For analysis of meta-models we propose to use Friedman's H-statistic~\cite{friedman2008predictive}. This method decomposes the prediction into components corresponding to two selected features. The variance of the difference between observed values and decomposed one without interactions is used to assess the strength of interactions. We use implementation of Friedman's H-statistic from \cite{molnar2018iml}.

\begin{figure}[h]
    \centering
    \includegraphics[width=0.9\textwidth ]{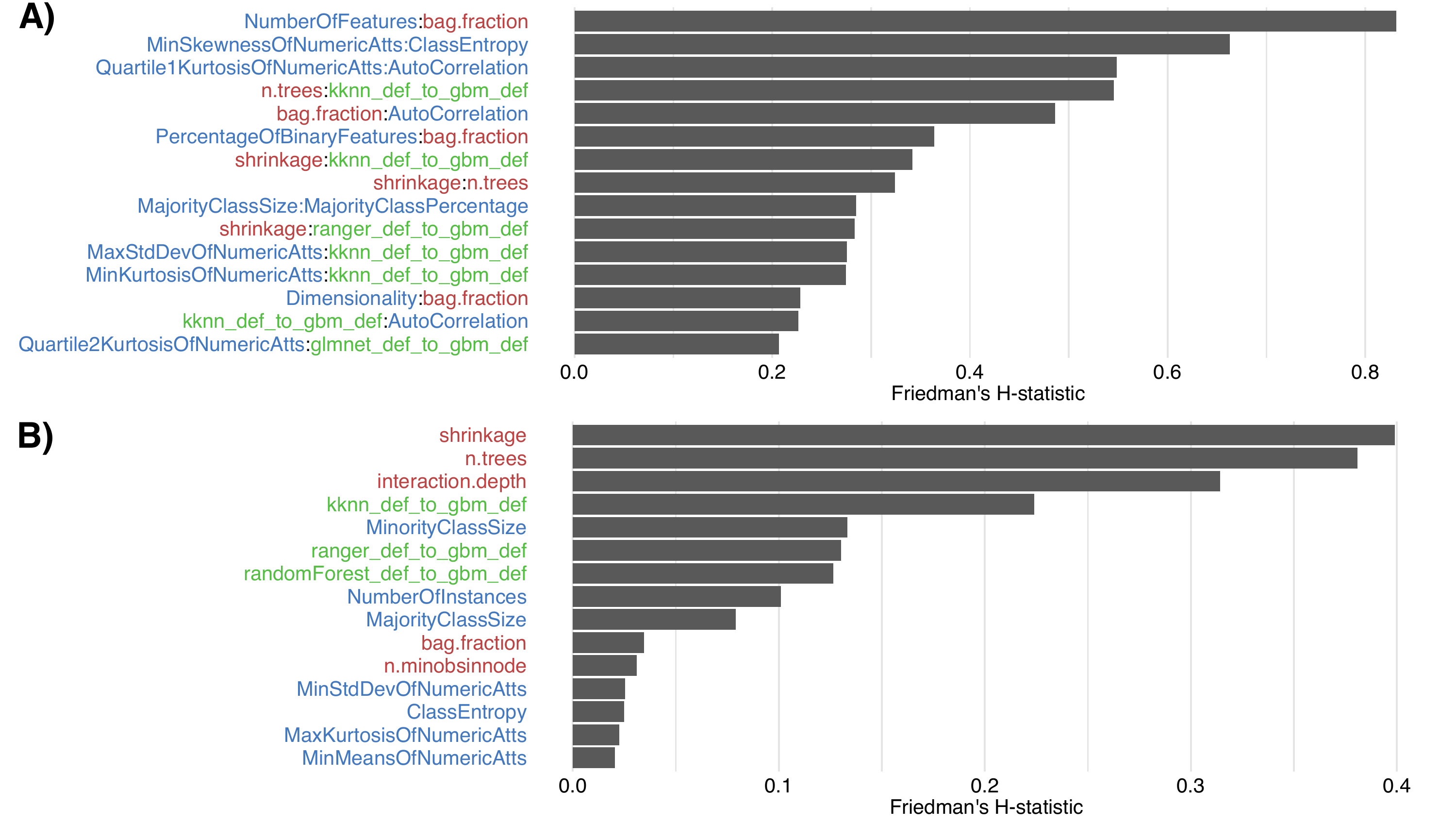}\vspace{-3mm}
    \caption{Panel A: Top  15  most  important two-way interactions. Panel B: Top  15  most  important meta-features in terms of overall propensity to interactions. Colors code groups of meta features}
   \label{two_way_interactions}

\end{figure}

The example identification of interactions for Meta-OpenMl100 is presented  in Figure \ref{two_way_interactions}. Firstly, we look at two-way interaction between any two meta-feature. The strongest interaction is for \texttt{bag.fraction} and \texttt{NumberOfFeatures}, i.e. between hyperparameter and statistical meta-feature. Based on such analysis we can directly identify which meta-charateristics are linked with selection of particular hyperparameters, which is a competitive approach to \cite{feurer2015initializing}.

In Figure~\ref{two_way_interactions}B there is the overall assessment of the variable propensity to interact with any other meta-feature. This approach is close to an extension of in \cite{van2018hyperparameter}. In this case, the ranking is similar to this at the Figure~\ref{plot_top15_most_imp_feature_agg}. The most prone to interact variables are hyperparameters.

\subsection{Importance of correlated meta-features}

Meta-features are often correlated because they describe similar characteristics of data sets. It is not known whether adding more correlated features increases the quality of the meta-model anyway.
To test the validity of a group of correlated variables we propose to use the triplot technique \cite{pekala2021triplot}.
Groups of meta-variables are determined according to hierarchical clustering based on correlations between meta-features. At each stage the importance of the selected meta-variables  cluster is calculated.

\begin{figure}[h]
    \centering
    \includegraphics[width=\textwidth ]{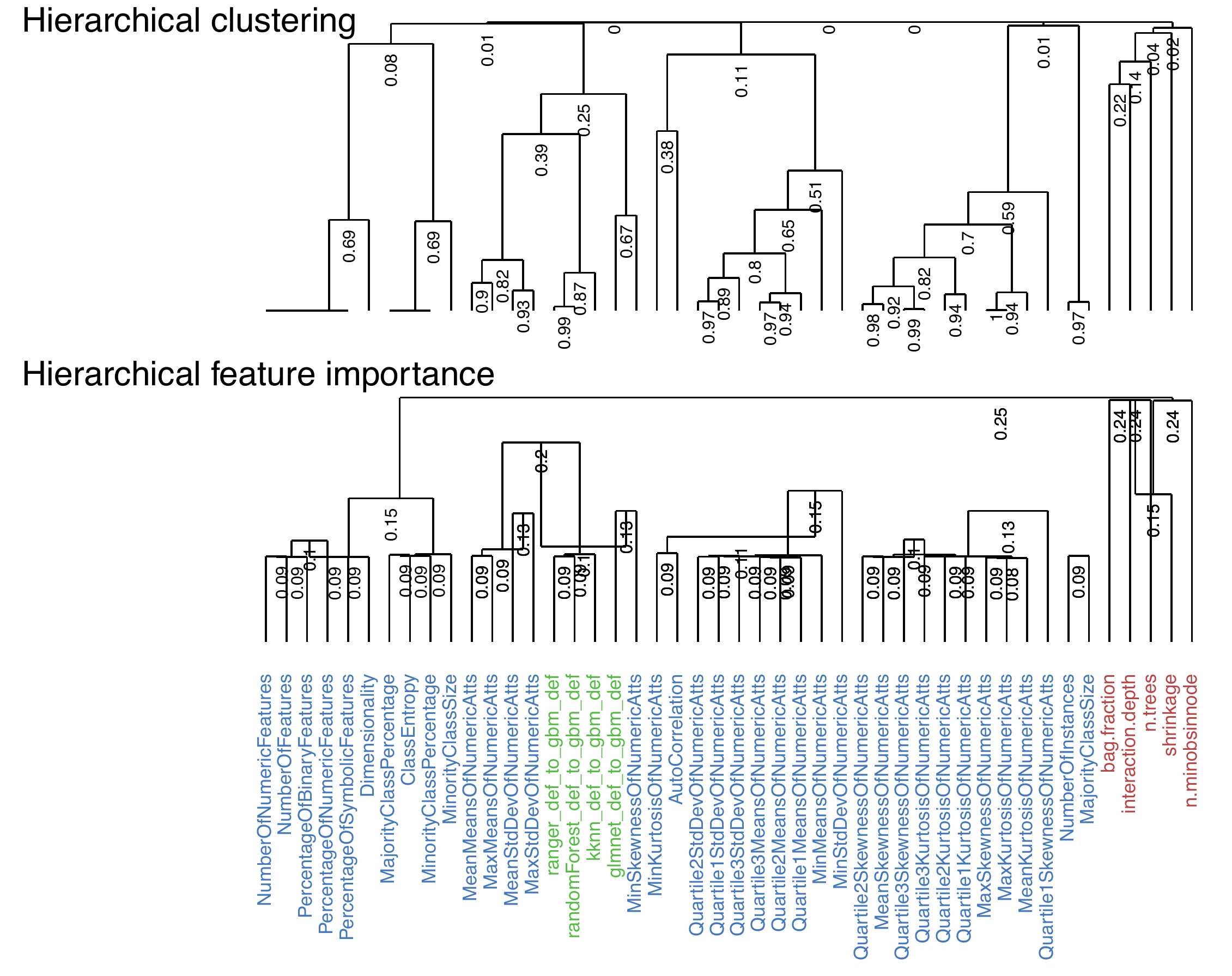}
    \caption{Triplot joint analysis of feature correlation and importance. The top panel shows hierarchical clustering for meta-features while the bottom one shows an importance of groups of features. Types of meta-features are color coded. }

    \label{hierarchical_cluster_importance}
\end{figure}

In Figure~\ref{hierarchical_cluster_importance} on the right panel we see that most of the statistical meta-features have marginal contributions independently and even after clustering their accumulated importance is substantially less than individual hyperparameters contributions. Landmarkers connected with information about mean values and standard deviation of numeric columns are very close to the importance of hyperparameters independently.

   \subsection{Hyperparameters informativeness}
   \label{sec:hyperparameters_informativeness}

   The majority of meta-learning is concentrated on obtaining the optimal hyperparameters configuration or effective warm-start points for the novel task. Some of them present the estimated empirical priors distribution of hyperparameters \cite{van2018hyperparameter}. Alternative approaches to estimating the learning profile for the selected hyperparameter are Ceteris Paribus (CP) profiles, known also as Individual Conditional Expectation profiles \cite{goldstein2015peeking}. This curve shows how a model’s prediction would change if the value of a single exploratory variable changed. In essence, a CP profile shows the dependence of the conditional expectation of the dependent variable (response) on the values of the particular explanatory variable. This is equivalent to a partial dependence plot for an individual instance.
      In this analysis, we use its R implementation  \cite{biecek2018dalex}.

 \begin{figure*}[h]
    \centering\includegraphics[height = 0.4\textwidth, width=\linewidth]{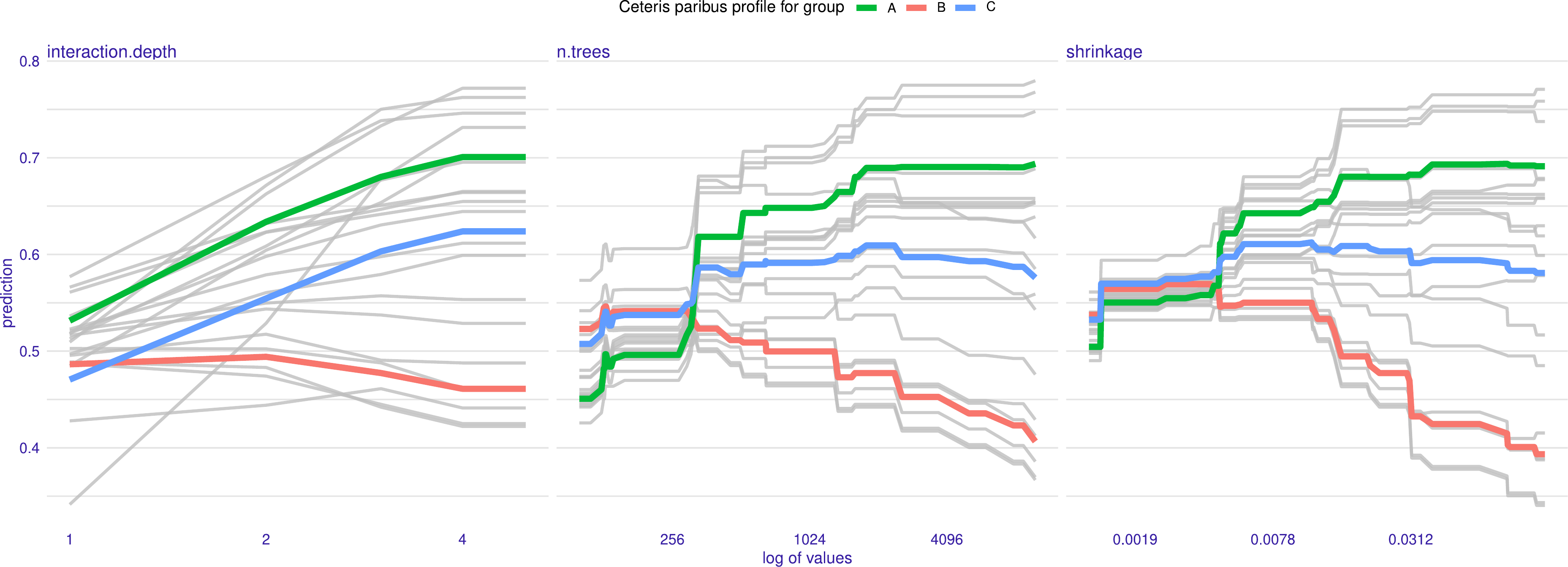}
\caption{Ceteris paribus for hyperparameters for test instances. Thick coloured lines are aggregated profiles for data sets clusters. Colours indicate groups. In the x-axis there is  applied logarithmic scale. }
\label{ceteris_paribus_hyperparamters_test}
 \end{figure*}

As we argue, looking at the profile, we would like to point the optimal hyperparameter value for independent data. Thus, we provide CP profile for every data set in meta-data  and  the curve is extracted from meta-model from the crossvalidation schema in which the selected task is the test example. 
In a result on each plot, we present CP profiles for every data set when the prediction curve is independent of that data. This situation is equivalent to the production application of meta-framework optimization.

Obtained CP analysis for the selected hyperparameters are presented in Figure \ref{ceteris_paribus_hyperparamters_test}.
To recall, the higher ranking, the better predictive power of the given for gradient boosting model. So observing increasing lines in CP profiles indicate the better rating of the considered gbm models.

.

  For \texttt{shrinkage}, \texttt{interaction.depth} and \texttt{n.trees} hyperparameters, we detect three patterns of profiles. We apply hierarchical clustering for profiles and the aggregated profiles for three groups are shown in Figure~\ref{plot_top15_most_imp_feature_agg}. These groups are indicated with distinctive colours and termed as  A, B and C. It is worth to highlight that the groups indicated as A for two hyperparameters may consist of different data sets because the clusterings are performed independently.
 
 For each of these three hyperparameters,  the group A has an increasing CP profile with a very strong trend for the interior values of variables. Profiles stabilise for high prediction for larger meta-features values and the maximal value of hyperparameters would be pointed as an optimal warm start point to optimization. Similar behaviour we can observe in group C.  Group B is strictly different for \texttt{n.trees} and \texttt{shrinkage}:  the highest prediction of rankings are obtained for the medium value of hyperparameters, and then predictions decrease.

 \subsection{Robustness of meta-data}
 
 In the case of the validation of a predictive model, it is important to check the influence of observations. This allows to avoid a situation in which one observation significantly influences the way the model works. In the case of meta-model, that influential meta-instance can disturb the selection of the optimal values of hyperparameters. Therefore, the similar analysis should be done for the meta-model. We propose to use two methods of calculating the observation influence, the Cook's distance and the distance between the estimated optimal hyperparameter values in a~\textit{full} meta-model (built on all meta-instances) and a~retrained meta-model omitting one data set.

For reliable validation, we select the \textit{full} model as one specified meta-model from the Meta-OpenMl100 one-data set-out schema: data set 1471 is the test instance. For this data, we check the change in optimal hyperparameter values. The influences of the remaining data sets are estimated by sequentially deletion them as we describe above.

\begin{figure}[h]
    \centering
            \includegraphics[width=0.9\textwidth]{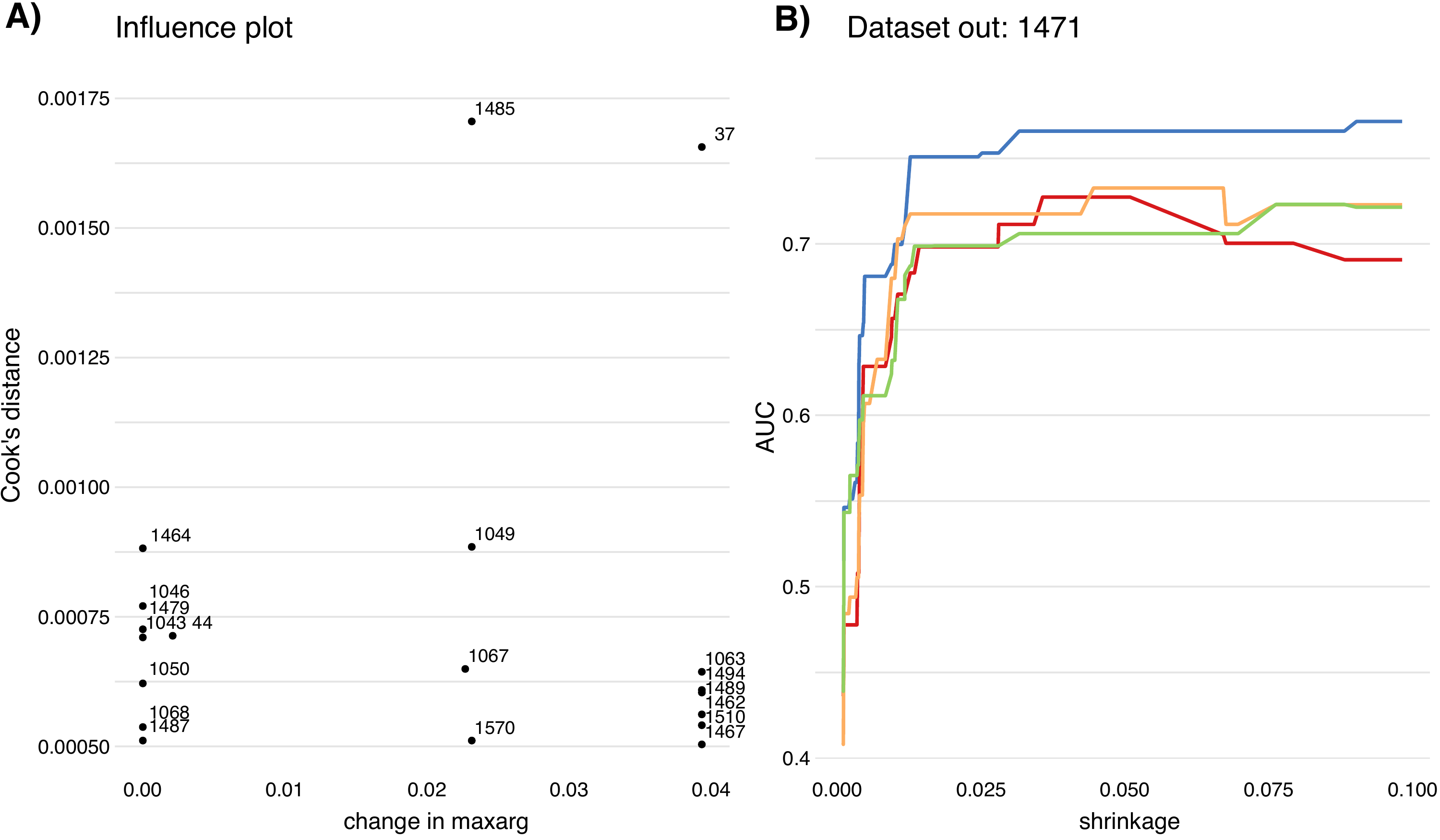}
      \caption{Panel A: Cook's distance vs. change in estimated optimal \texttt{shrinkage}. Numbers stand for OpenML data sets ids. Panel B: Individual profiles for the most and the least influential data set.   Blue line corresponds to the \textit{full} model. } 
      \label{plots_influential}

\end{figure}

In Figure~\ref{plots_influential}A we see the scatter plot for the considered measures of disturbance caused by removing the effect of the single data set. On the x-axis there is distance between the optimal shrinkage hyperparameter, on the y-axis there is the value of Cook's distance. We see that the biggest perturbations in the final predictions are caused by deleting the data sets $1485$ and $37$. For the data set $37$ we also observe the most significant shift of optimal hyperparameter value. 

In Figure~\ref{plots_influential}B are CP profiles for the full meta-model and for the selected limited meta-models with diverse values of Cook's distance. These profiles allow us to assess the significance of the change in the choice of the optimal point. Only  for the data set $37$ we observe the transformation of CP profile in comparison to the~\textit{full} model. This observation shows that the selection of meta-data in this particular example provides the robust selection of hyperparameters warm-start.

\section{Conclusions}

Meta-learning is a promising approach for AutoML solutions. The main contribution of this work is to show the applications of techniques known in explainable artificial intelligence area in meta-learning. It turns out that many model exploration techniques not only increase our knowledge of which meta-features are relevant, but also increase our knowledge of the relationship of hyperparameters to model performance. In this work, we have shown how to use XAI techniques to not only build an effective meta-model, but also to extract knowledge about the importance of the particular meta-features in the meta-model. Variable importance technique and extension of this taking into account correlation structure of meta-feature amplify the selection of the most informative set of meta-features. This is crucial in the trial-and-error process of defining meta-space and brings us closer to find transferable representation of data sets. Examining interaction between data set based meta-features and hyperparameters configuration along with Individual  Conditional  Expectation profiles for hyperparameters supports hyperparameter tuning. In further research, this may lead to a significant reduction of the dimension of the searched hyperparameter space and an improvement of automatic model selection processes. The quality of meta-features and evaluation of hyperparameters is determined by the choice of datasets. Validating data sets informativeness through Cook's distance helps to build a robust and reliable repository for meta-learning.

This approach is universal and generic to the explainable analysis of any meta-learning model presented in Figure \ref{graphical_abstract}.  The OpenML may be argued that this is not the appropriate illustration for every meta-learning problem - the datasets are relatively small. However, at the moment there are no publicly available repositories for big data problems. This approach can be reproduced on any repository of data sets from a specific domain, or data sets of a selected size or complexity. It is sufficient to store a selected set of meta-targets, fixed hyperparameters and models performance, and then build a black box meta-model to predict model performance for novel data. 

%apalike
\bibliographystyle{splncs04}
\bibliography{references}
\end{document}